# Can Coding Agents Reproduce Official Statistics? Metadata, Retry Budget and the Limits of Execution Feedback in a Controlled Eurostat Benchmark

**Sabina-Cristiana Necula**
Alexandru Ioan Cuza University of Iași, Faculty of Economics and Business Administration
Iași, Romania
*sabina.necula@uaic.ro*

**ABSTRACT**

*Large language models can generate executable data-analysis code, but successful execution is not equivalent to a valid official-statistics result. This study asks whether authoritative metadata and execution feedback improve the reproducibility of Eurostat answers produced by a coding agent, and isolates what execution feedback actually contributes. A benchmark of 30 natural-language tasks covering seven domains, seven Eurostat datasets and four difficulty tiers was run under four conditions: task only (A), task plus a frozen dataset metadata card (B), metadata plus a repair loop driven by sanitized execution feedback (C), and metadata plus the same attempt budget with no diagnostics of any kind (D). Claude Sonnet 5 generated Python through the Anthropic Messages API in three independent replicates, yielding 360 task-runs. Exact correctness required successful execution, the correct dataset, filters, output shape, values and unit. Metadata raised exact correctness from 27.8% to 51.1% (+23.3 points, 95% task-cluster bootstrap CI 8.9 to 37.8; Holm-adjusted exact McNemar p=0.0049) and almost eliminated wrong-filter outcomes. Both C and D reached 76.7%, each exceeding B by 25.6 points (p=0.0013 and p=0.00047), but they did not differ from one another: the paired difference was exactly 0.0 points (95% CI -12.2 to 13.3; p=1.000), with twelve discordant pairs in each direction. Sanitized diagnostics therefore contributed nothing beyond the retry budget they accompany. A companion experiment run under an under-specified output contract, in which the required ranking key and unit representation were never stated to the model, understated condition C by 23.4 points, showing that evaluator and contract design can dominate measured agent error. Reliable statistical coding agents need semantic validation against frozen specifications, a fully specified output contract, and a retry budget — not execution diagnostics.*

**Keywords: large language models; code generation; official statistics; Eurostat; execution feedback; retry budget; output contract; reproducibility**
**JEL Classification:** C82; C88; C63

## 1. INTRODUCTION

Natural-language interfaces to official statistics promise to shorten the path from a question to a reproducible result. A user may ask for a population value, an unemployment-rate gap or a ranking of countries and receive Python that queries an official application programming interface (API), transforms the response and returns a machine-readable answer. The workflow is attractive because the generated program can, in principle, be inspected and rerun. Yet the same workflow creates a distinctive reliability problem: a program may execute successfully while selecting the wrong indicator, population, unit, time range or direction of comparison. In official statistics, such silent semantic errors are more consequential than a visible runtime exception because a plausible number can pass into analysis without triggering a warning.

The European Statistics Code of Practice places accuracy, reliability and systematic validation at the centre of statistical quality (European Commission, 2017). Those requirements do not disappear when an LLM writes the retrieval code. Standard code-generation benchmarks usually assess whether programs satisfy tests or return an expected output (Chen et al., 2021; Lai et al., 2023). Eurostat tasks add a semantic layer: dimensions are encoded with compact labels, several measures coexist in the same dataset, and changes may be expressed as percentages or percentage points. A complete evaluation must therefore distinguish syntactic validity, operational success and statistical validity.

This paper reports a controlled benchmark of one coding agent under four information regimes. Condition A supplied only the task and the required output contract; B added a frozen metadata card for the named dataset; C added an iterative repair mechanism driven by sanitized execution feedback; and D granted the same attempt budget as C while withholding every diagnostic, so that a failed attempt was simply redrawn. The study asks four questions. First, does authoritative metadata raise exact statistical correctness? Second, does an execution-repair loop improve executability and exact correctness beyond metadata alone? Third — the question conditions A to C cannot answer, because feedback and attempt budget move together — is the benefit of that loop attributable to the diagnostics or merely to the additional attempts? Fourth, what errors remain when generated programs run? The contribution is an official-statistics-specific benchmark that couples frozen Eurostat snapshots with component-level scoring, paired repeated runs, a resampling control that identifies the effect of feedback, and an explicit audit of how much measured error is produced by the output contract itself rather than by the model.

## 2. RELATED WORK AND EXPECTATIONS

### 2.1. Executable code is an incomplete validity criterion

HumanEval established execution-based functional correctness as a central measure for code-generating language models (Chen et al., 2021). DS-1000 extended evaluation to realistic data-science problems and demonstrated the value of combining functional tests with surface constraints to reduce false acceptance (Lai et al., 2023). Text-to-SQL research similarly shows that database-grounded tasks require generalisation across schemas and complex query structures; Spider was designed precisely to separate linguistic interpretation from memorisation of database-specific programs (Yu et al., 2018). These benchmarks motivate executable evaluation but do not fully capture the statistical meaning of a result returned from a multidimensional dissemination system.

Eurostat's Statistics API is a public REST service that provides JSON-stat 2.0 data (Eurostat, 2026a). A correct program must find out the category codes within ordered dimensions, re-build the intended slice, manage missing values and keep the requested unit. A malformed flat-index calculation may still return a number. Again, using total population instead of the labor force, nominal GDP instead of real GDP or an index instead of a rate can produce syntactically valid output using the wrong estimand. Thus the benchmark considers execution success a secondary operational outcome. The benchmark specifies exact correctness as six simultaneous checks.

### 2.2. Metadata and execution-guided repair

Schema information can be used to refine the mapping from a natural-language request to data dimensions. Here the metadata card reports the size of the dataset, the relevant codes and meanings of units, but not the gold value. This intervention is similar to providing database schema context in semantic parsing, but it also encodes statistical concepts such as age coverage, denominator and national-accounts item. The first expectation, then, is that B will reduce filter and unit selection error relative to A.

Execution feedback is another kind of signal. Self-debugging methods can improve code generation if the model can see test or runtime information (Chen et al., 2024), but the gains reported depend on the quality of the feedback and the tests available. Olausson et al. (2024) show that self-repair gains are often modest once repair cost is taken into account, and that weak self-feedback is a bottleneck. Their analysis also cautions of a repair loop that bundles two interventions, as the model receiving feedback also receives another attempt. In a query of official statistics, a runtime error informs about code mechanics but doesn't generally tell that a plausible value corresponds to the wrong statistical concept. We then expect C to increase execution success more than exact correctness, that silent semantic errors will remain common after repair, and that a resampling control, given the same attempt budget but no diagnostics, will recover an appreciable fraction of the gains from C.

## 3. DATA AND METHOD

### 3.1. Benchmark construction

The benchmark contains 30 English-language requests distributed across population, labour market, national accounts, prices, energy, education and labour, and the digital economy. The requests use seven Eurostat datasets: tps00001, une_rt_a, nama_10_gdp, prc_hicp_aind, sdg_07_40, lfsi_neet_a and isoc_eb_ai. Every task names its dataset so the experiment measures code construction and statistical selection rather than open-ended dataset discovery. Outputs are scalars, time series or rankings.

Tasks were assigned to four difficulty tiers. Tier 1 covers direct retrieval (seven tasks); tier 2 covers short series or a direct rate (two tasks); tier 3 covers transformations, differences, averages and rankings (17 tasks); and tier 4 covers concept-sensitive selections (four tasks). Examples include female-minus-male unemployment gaps, current-price GDP, household consumption as a share of GDP and AI use by enterprise size. Each task specifies a numerical tolerance appropriate to the published unit.

Gold answers were generated programmatically from seven frozen JSON-stat snapshots captured on 31 August 2026. The snapshots and task specifications were fixed before measured calls and were never sent to the model. This design protects the target values from prompt leakage and avoids score changes caused by later Eurostat revisions. Because the generated programs query the live dissemination API rather than a replay of the snapshot, an alignment audit accompanies every run: for each task it verifies that the live slice returned during execution equalled the frozen snapshot. All 30 tasks are confirmed in both experiments reported below, so no gold value drifted between capture and execution. The prompt shown in each condition included only the current task; B, C and D received only the metadata card for the task's dataset.

### 3.2. Experimental conditions and model configuration

Table 1 summarises the controlled interventions. In A, the model received the natural-language request, the named dataset and the output contract. B appended an authoritative, frozen card describing the current dataset's dimensions and codes. C used the same initial prompt as B; if static analysis, execution or output-schema validation failed, the model received sanitized diagnostics and could make up to two repairs, three attempts in total. It never received a gold value, a correct program or semantic correctness feedback. D used the same prompt and the same three-attempt budget as C, but a failed attempt was answered by resending the original request unchanged, with no diagnostics, no previous program and no indication that anything had gone wrong. The contrast between C and D is the identifying comparison of this study: it holds information and attempt budget constant and varies only whether the model is told what went wrong.

**Table 1.** Experimental conditions

| Condition | Information available | Execution feedback | Maximum attempts |
|---|---|---|---|
| A | Task, named dataset and output contract | None | 1 |
| B | A + frozen metadata card for current dataset | None | 1 |
| C | Same initial information as B | Sanitized static, runtime or schema diagnostics | 3 |

| Conditi on | Information available | Execution feedback | Maximum attempts |
|---|---|---|---|
| D | Same initial information as B | None; the request is resent unchanged | 3 |

*Note: condition C received no gold values and no semantic-correctness feedback.*

All measured calls used Anthropic Claude Sonnet 5 through the native Messages API and Structured Outputs (Anthropic, 2026a). The response schema required three fields: Python code, the asserted Eurostat dataset code and a short rationale. The output budget was 8,192 tokens; thinking was disabled; and no temperature, top-p or top-k value was set. Structured Outputs constrained the API response envelope, but the generated program still had to emit a separate JSON object after execution. The model, prompt files, metadata cards, task set and gold snapshots were held constant across three independent replicates completed on 2 September 2026. The order of conditions and the order of tasks within each condition were randomised per replicate from a recorded seed, so that condition is not confounded with time of execution; the realised order is archived in a run manifest. The resulting design comprises 30 tasks x four conditions x three replicates = 360 task-runs.

Generated code was treated as untrusted. A static checker restricted imports and dangerous constructs, network access was limited to the official Eurostat endpoint, and each accepted program ran in an isolated subprocess with a 45-second timeout and without the parent API credential. The execution contract required a single JSON object containing task ID, dataset code, filters, result, unit and notes. These controls reduce risk and standardise evaluation; they are not a claim that language-model-generated code is safe for unrestricted production use.

### 3.3. Outcomes and statistical analysis

The primary outcome, exact correctness, is obtained when all six components are satisfied: the program runs and produces valid output; the dataset is correct; substantive filters are correct; the shape of the result is correct; numerical values are within tolerance of the frozen gold answer; and the unit is semantically correct. Execution logs the first requirement only for success. A silent error is an executable, schema-valid output that fails at least one of the other exactness components. Errors were classified into a mutually exclusive taxonomy based on the first semantic component that failed after operational checks.

Proportions are given with 95% Wilson confidence intervals (Wilson, 1927). Since each task was performed under all conditions, discordant paired outcomes were compared with exact two-sided McNemar tests (McNemar, 1947); the three pairwise p-values for each outcome were corrected with Holm's sequential procedure (Holm, 1979). There are 90 task-replicate pairs nested within 30 tasks, so paired differences are also accompanied by 95% task-cluster bootstrap intervals from 50,000 resamples of tasks, using seed 20260831. Descriptive are the tier and domain analyzes given the smallness of several of the strata.

### 3.4. The output contract as an experimental factor

The benchmark was first run on 1 September 2026 under an output contract that, in hindsight, was incomplete. The contract required a printed JSON object with six named keys and instructed the program to use Eurostat codes in its filters and result labels, but it never stated two things the evaluator nevertheless required: the key under which a ranking entry should carry its dimension code, and whether the unit field should hold the official Eurostat code or its documented label. Generated programs consequently reported rankings as {"geo": "RO", "value": ...} and units as CP_MEUR or PC_GDP, both of which satisfy everything the contract stated, while the evaluator demanded the literal key "code" and the labels million euro and percent of GDP. Answers whose dataset, filters, ordering and figures matched the frozen gold exactly were therefore scored as wrong.

Rather than silently rescore that experiment, the study treats the completeness of the contract as an experimental factor and reports two runs. The first, under the original contract, is retained unchanged. The second, reported as the primary experiment throughout Section 4, uses a contract that states the required shape of the result for each result shape — including the ranking key — and the required form of the unit, specifying that a difference between categories or a change between periods is reported in percentage points rather than in the level unit. Nothing else differs: the same tasks, gold snapshots, metadata cards, model, sampling settings and scoring code. Section 4.3 quantifies the gap between the two, which is a direct measurement of how much apparent model error an under-specified contract can manufacture.

The scoring code makes both readings reproducible rather than asking the reader to trust a judgement. Unit equivalence is resolved through the unit dimension published with the seven benchmark datasets, so that an official code and its documented meaning are one unit; the rule is deliberately asymmetric, because a code that names the unit of the stored series is accepted only for operations that preserve that unit, and percentage and percentage-point answers are never merged. The original literal comparison and a stricter, denominator-preserving rule are retained as switches, as is the ranking-key rule. Each produces its own result directory from the same raw records. No generated code, response, filter, value, shape or execution output was ever altered, and no run was repeated to obtain a better score.

## 4. RESULTS

### 4.1. Primary outcomes

All 360 task-runs of the primary experiment completed the API and schema stage except one, a condition D run retained as an Anthropic InternalServerError. The primary results are shown in Table 2 and Figure 1. Exact correctness was 25/90 (27.8%) in A, 46/90 (51.1%) in B, and 69/90 (76.7%) in both C and D. The task-clustered differences were 23.3 percentage points for B minus A (95% bootstrap CI 8.9 to 37.8), 25.6 points for C minus B (12.2 to 37.8) and 25.6 points for D minus B (14.4 to 36.7). The difference between C and D was exactly zero (-12.2 to 13.3).

Paired exact McNemar tests lead to the same interpretation (Table 3). Metadata improved exact correctness over task-only prompting (Holm-adjusted p=0.0049). Both C and D exceeded B (p=0.0013 and p=0.00047) and A (p<0.0001). The comparison of C with D, however, produced twelve discordant pairs in each direction out of 90, an exact p of 1.000 and a task-clustered interval that is symmetric about zero. The two conditions did not merely fail to differ significantly; they returned the identical count of exact answers while differing only in whether the model was told what had gone wrong.

Execution success followed a related pattern. A executed successfully in 71/90 cases (78.9%), B in 69/90 (76.7%), C in 90/90 (100.0%) and D in 87/90 (96.7%). Both repair conditions exceeded A and B by wide margins, while C exceeded D by 3.3 points with an interval touching zero (-7.8 to 0.0). Whatever advantage diagnostics confer is therefore confined to the last few programs that fail to run at all, and does not propagate to statistical validity.

**Table 2.** Primary outcomes by condition

| Condition | Execution success | Exact correctness (95% CI) | Silent error among executable | Mean attempts |
|---|---|---|---|---|
| A | 71/90 (78.9%) | 25/90 (27.8%; 19.6–37.8) | 46/71 (64.8%) | 1.000 |
| B | 69/90 (76.7%) | 46/90 (51.1%; 40.9–61.2) | 23/69 (33.3%) | 1.000 |
| C | 90/90 (100.0%) | 69/90 (76.7%; 67.0–84.2) | 21/90 (23.3%) | 1.233 |
| D | 87/90 (96.7%) | 69/90 (76.7%; 67.0–84.2) | 18/87 (20.7%) | 1.189 |

*Note: Wilson 95% confidence intervals are shown for exact correctness; the silent-error denominator is executable outputs. Conditions C and D share the same three-attempt budget.*

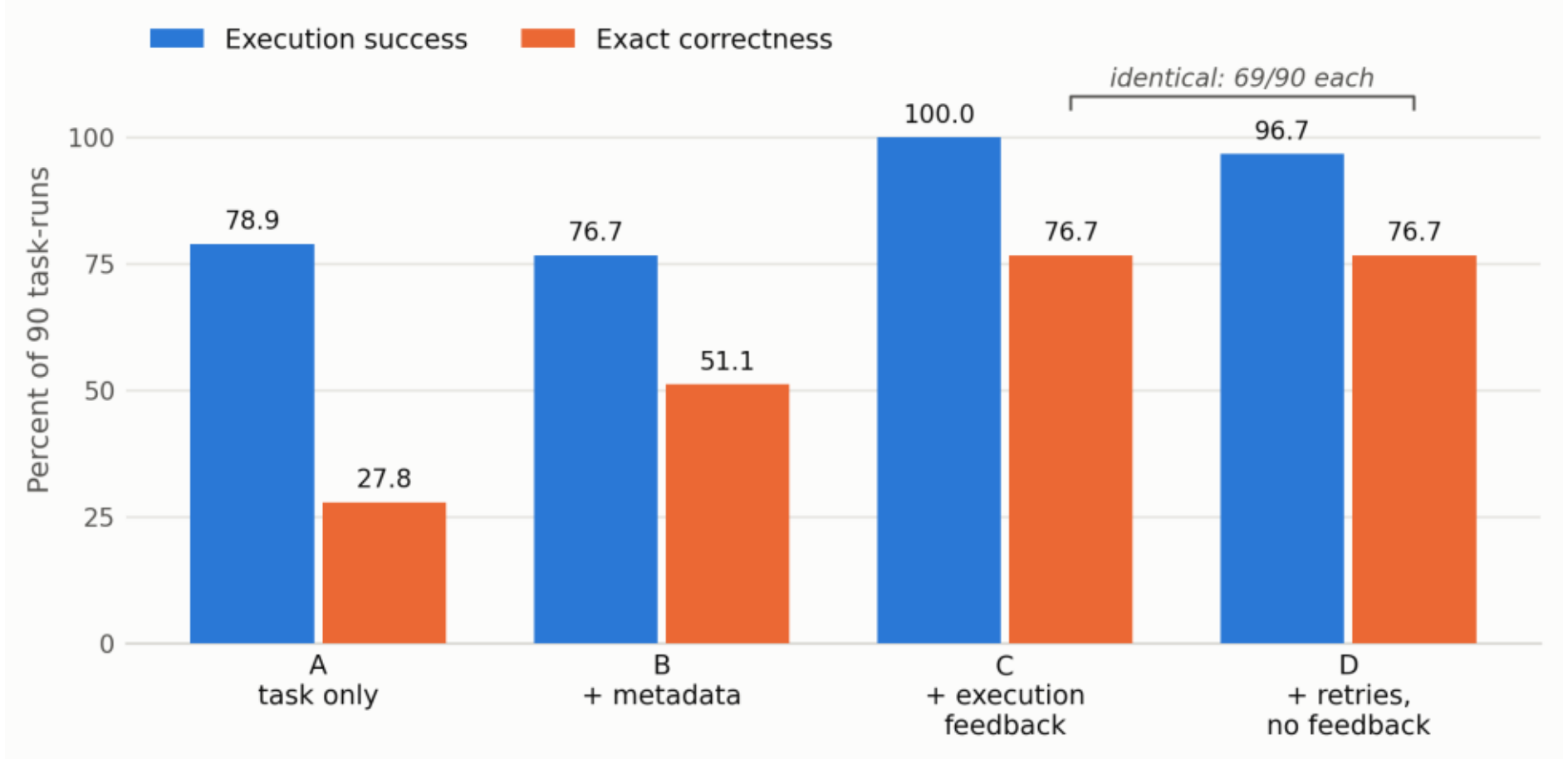

**Figure 1. Execution success and exact correctness across the four conditions (n=90 task-runs per condition, specified output contract)**

**Table 3.** Paired comparisons of exact correctness

| Comparison | First only correct | Second only correct | Difference, pp (cluster 95% CI) | Holm-adjusted McNemar p |
|---|---|---|---|---|
| A vs B | 12 | 33 | +23.3 (8.9 to 37.8) | 0.0049 |
| A vs C | 7 | 51 | +48.9 (30.0 to 65.6) | <0.0001 |
| A vs D | 5 | 49 | +48.9 (33.3 to 64.4) | <0.0001 |
| B vs C | 9 | 32 | +25.6 (12.2 to 37.8) | 0.0013 |
| B vs D | 6 | 29 | +25.6 (14.4 to 36.7) | 0.0005 |
| C vs D | 12 | 12 | +0.0 (-12.2 to 13.3) | 1.0000 |

*Note: difference is second minus first. Cluster intervals resample 30 tasks and retain all three replicates. Holm adjustment is applied across the six comparisons.*

## 4.2. Silent errors, repair and error composition

Among executable outputs, the silent-error rate fell from 46/71 (64.8%) in A to 23/69 (33.3%) in B, 21/90 (23.3%) in C and 18/87 (20.7%) in D. The ordering matters less than the level: even under the best condition, one executable answer in five was silently wrong, so a green execution signal remains an unreliable proxy for statistical validity.

The process of the retry effect is demonstrated in the tries themselves. Condition C triggered a repair on 20 of its 90 task-runs; all 20 became executable and 14 became exactly correct. Condition D redrew on 15 task-runs, 13 of which became executable and 10 of which became exactly correct. Thus, the conversion rate from a retried run to an exact answer is 70% with diagnostics and 67% without them. At the level of individual runs the null result for C minus D means that a second draw from the same distribution fixes a failing program about as often as a diagnosed repair does. D reached its outcome with a slightly lower mean number of attempts than C, 1.189 vs 1.233 which is worth noting. Table 4 shows where the remaining errors are. Condition A resulted in 39 wrong-filter outcomes, while there were four in B, two in C and three in D. This reduction is consistent with metadata supplying the missing dimension codes and reproduces the clearest finding of the earlier experiment. Once filter selection improved, wrong shapes and wrong units became the dominant residual semantic errors, at nine and six in C and six and seven in D. Two further categories are properties of the harness rather than of statistical reasoning: 15 outcomes in B and two in D were terminal rejections by the static checker, in every case for importing sys, and 19 outcomes in A were runtime or schema failures, of which 13 were programs that printed something other than a single JSON object.

**Table 4.** Mutually exclusive outcome taxonomy

| Outcome | A | B | C | D |
|---|---|---|---|---|
| Exact correct | 25 | 46 | 69 | 69 |
| Wrong filters | 39 | 4 | 2 | 3 |
| Wrong shape | 4 | 10 | 9 | 6 |

| Outcome | A | B | C | D |
|---|---|---|---|---|
| Wrong unit | 2 | 7 | 6 | 7 |
| Wrong values | 1 | 2 | 4 | 2 |
| Safety rejection | 0 | 15 | 0 | 2 |
| Runtime or schema failure | 19 | 6 | 0 | 0 |
| API or schema failure | 0 | 0 | 0 | 1 |
| Total | 90 | 90 | 90 | 90 |

*Source: primary experiment under the specified output contract; pilot records excluded.*

### 4.3. Difficulty, domains, and the cost of an under-specified contract

Difficulty results are exploratory but informative (Table 5). Transformation and ranking tasks, which make up 17 of the 30 tasks, reached 73% under C and 75% under D, against 29% under A. Concept-sensitive tier-4 tasks reached 67% under C and 92% under D, though with only twelve task-runs per condition that difference should not be read as a ranking. The domain pattern was heterogeneous: under C the labour-market tasks reached 15 of 15 exact outcomes while price tasks reached 7 of 12. Each domain contains only 12 or 15 task-runs, so these differences should not be treated as stable estimates. Because tier 3 contains more than half the benchmark, the headline exact rate is substantially determined by tier composition, and a benchmark weighted towards direct retrieval would report a higher figure for the same system.

The companion experiment quantifies what the output contract itself costs (Table 6). Under the original, under-specified contract the same model, tasks and scoring code produced 23/90 (25.6%) in A, 40/90 (44.4%) in B and 48/90 (53.3%) in C. Specifying the ranking key and the unit representation raised those to 27.8%, 51.1% and 76.7%. The effect is concentrated exactly where the specification was missing: condition C gains 23.4 points, and the seven ranking tasks account for most of it.

Two independent arguments lead to the same conclusion. Rescoring the original run using a rule that accepts a ranking's dimension code under any key returns 74.4% for condition C, within two points of the 76.7% obtained by specifying the contract and rerunning. So the difference was one of serialization, not of statistical capability. The original figures were twenty points or more below the agent. In the original contract the comparison of C with B was not significant (adjusted $p=0.2153$); in the rescoring it was ($p=0.0237$); and in the specified contract it is ($p=0.0013$). A conclusion about execution feedback therefore in the first instance depended on a JSON key name that was never communicated to the model. Reporting both experiments makes the dependence visible rather than resolving it by fiat.

**Table 5.** Exact correctness by difficulty tier

| Tier | Description | Task-runs per condition | A | B | C | D |
|---|---|---|---|---|---|---|
| 1 | Direct retrieval | 21 | 9 (42.9%) | 12 (57.1%) | 19 (90.5%) | 17 (81.0%) |
| 2 | Series/direct rate | 6 | 1 (16.7%) | 4 (66.7%) | 5 (83.3%) | 3 (50.0%) |
| 3 | Transformation/ranking | 51 | 15 (29.4%) | 25 (49.0%) | 37 (72.5%) | 38 (74.5%) |
| 4 | Concept-sensitive | 12 | 0 (0.0%) | 5 (41.7%) | 8 (66.7%) | 11 (91.7%) |

*Note: subgroup results are descriptive; denominators are small.*

**Table 6.** Exact correctness under the original and the specified output contract

| Condition | Original contract | Specified contract | Difference |
|---|---|---|---|
| A | 23/90 (25.6%) | 25/90 (27.8%) | +2.2 pp |
| B | 40/90 (44.4%) | 46/90 (51.1%) | +6.7 pp |
| C | 48/90 (53.3%) | 69/90 (76.7%) | +23.4 pp |
| D | not run | 69/90 (76.7%) | — |

*Note: both experiments use the same tasks, gold snapshots, metadata cards, model settings and scoring code; they differ only in whether the output contract states the ranking key and the unit representation. Condition D exists only in the specified-contract experiment.*

## 5. DISCUSSION

The experiment pulls apart three things that are usually packaged together in agentic systems: the information the model is fed, the number of tries it gets, and the diagnostics it receives when a try fails. Metadata changed what the program requested Eurostat to return. The attempt budget changed the frequency of getting a usable program at all. Holding the budget constant, the diagnostics made no measurable difference. The increase of 23.3 points from A to B, combined with the reduction of wrong-filter outcomes from 39 to four, shows that compact authoritative context solves a large fraction of statistical selection failures, and that providing the relevant statistical ontology at point of generation is a stronger intervention than asking the model to reason longer.

The study's sharpest result is the comparison of C with D. Both conditions had the same metadata and the same three attempts; only C was told what had gone wrong. They came up with the same 69 responses out of 90, with 12 pairs that were discordant in either direction, and D used slightly fewer tries to get there. In combination with the retried runs, 14 of 20 converted to exact answers in C, and 10 of 15 in D. The finding is not that feedback is weakly helpful, but that it is doing the work that a second sample would have done anyway in this setting. This expands the caution of the self-repair literature (Chen et al., 2024; Olausson et al., 2024) with an explicit control: earlier work compares repair to a single attempt, which fails to separate the diagnostic from the draw.

The interpretation should be bounded in three ways. The interval for C minus D runs from -12.2 to 13.3 points, so a small effect cannot be excluded, only one large enough to matter operationally. The diagnostics here are deliberately thin — sanitized static, runtime and schema messages, with no semantic content — so the result speaks to operational feedback, not to richer validation signals, which would constitute a different intervention and would risk leaking part of the target specification. Finally, a residual advantage of C over D survives on executability alone, 100% against 96.7%, so diagnostics do appear to rescue the last few programs that cannot be made to run by redrawing.

For official-statistics systems the practical architecture follows directly, and it is cheaper than the one a naive reading of the repair literature would suggest. Retrieval should be grounded in versioned metadata for the selected dataset. The output contract should be stated in full, including the serialisation of every result shape and of the unit, because anything left implicit will be scored as model error. Code should run in a constrained environment with an explicit output contract and a retry budget, which is the component that actually buys reliability. A semantic validator should then compare dataset, dimension codes, shape, unit and calculation logic against a machine-readable task specification, and the system should abstain or escalate when that check is unavailable. Engineering effort spent on sanitized diagnostics is, on this evidence, better spent on the validator.

The cost result is encouraging but should be interpreted narrowly. The primary experiment consumed 397 API calls, including retries, for 688,966 input and 510,011 output tokens. At the provider's list price of USD 2 per million input tokens and USD 10 per million output tokens (Anthropic, 2026b), direct API cost was approximately USD 6.48 for 360 task-runs. This excludes benchmark construction, engineering, manual audit and the institutional cost of a wrong published statistic. The marginal inference cost is therefore small relative to the governance and validation burden, and small enough that a resampling budget is an inexpensive reliability mechanism.

The findings also matter for benchmark design. DS-1000 emphasises reliable acceptance tests because an evaluator that accepts incorrect code can invert model comparisons (Lai et al., 2023). The present study illustrates the complementary failure and measures it: an evaluator that requires a serialisation the prompt never specified rejects correct answers, and here that false rejection was large enough to reverse the sign of a reported conclusion. Official-statistics benchmarks should therefore publish the full output contract shown to the model alongside the scoring code, dimension mappings, frozen source responses, scoring precedence and sensitivity results, so that both false acceptance and false rejection can be examined.

## 6. LIMITATIONS AND VALIDITY

Several limitations bound the conclusions. First, the study evaluates one proprietary model version on two dates. Model updates, default behaviours and safety systems can change, so the observed rates are not a permanent ranking of Claude or of LLMs generally. Second, the 30 tasks cover seven datasets and deliberately name the dataset. Open-ended dataset discovery, conversational clarification, multilingual requests and other dissemination systems remain outside scope.

Third, the three replicates provide stochastic repetition but only 30 independent task clusters. Cluster bootstrap intervals address the repeated-task structure, yet subgroup estimates are imprecise and no interaction should be inferred from the tier or domain tables. Fourth, the null result for C minus D is a null within a bounded interval, not a proof of equivalence, and it applies to sanitized operational diagnostics; a feedback channel carrying semantic content might behave differently.

Fifth, the two experiments differ in the completeness of the output contract, which is the intended manipulation, but they were also run on consecutive days and the second randomised condition and task order while the first did not. The contract effect is therefore estimated between experiments rather than within a single randomised design, and the alignment audit rather than randomisation is what rules out data drift. Sixth, terminal rejections by the static checker are counted as failures, which attributes part of the between-condition difference to the benchmark harness rather than to the model; the taxonomy reports them separately so that this can be inspected. Finally, the static checker and subprocess isolate common hazards but are not a general security sandbox, and production deployment requires system-level network and filesystem controls.

## 7. CONCLUSION

Execution alone was an unreliable proxy for statistical correctness: even in the best condition, one executable answer in five was silently wrong. Authoritative metadata raised exact correctness from 27.8% to 51.1% and almost eliminated wrong-filter outcomes. Adding a three-attempt budget raised it further to 76.7%, but a control that received the same budget with no diagnostics whatsoever reached exactly the same 76.7%, so the contribution of sanitized execution feedback beyond the retries it accompanies was not detectable. A companion experiment showed that an under-specified output contract had understated the same system by more than twenty points and had reversed a reported significance test. Trustworthy natural-language access to official statistics therefore rests on three things that are cheaper than a repair loop: metadata grounding, a fully specified output contract, and independent semantic validation — with a retry budget, rather than diagnostics, as the mechanism that makes generated programs usable.

## DATA AND CODE AVAILABILITY

The replication package is archived on Zenodo at https://doi.org/10.5281/zenodo.22253054 (the version used for this paper, deposited 2 September 2026); https://doi.org/10.5281/zenodo.22253053 always resolves to the latest version. Code is released under the MIT Licence, and the data, result tables and documentation under CC BY 4.0; the frozen Eurostat JSON-stat snapshots are redistributed unmodified under Eurostat's reuse policy, with Eurostat as the source. The package contains the task registry, both output contracts, metadata cards, frozen snapshots, gold specifications, generated programs, raw records, execution outputs, evaluation scripts, the run manifest with the randomisation seeds, and derived tables. The prompts used in the first experiment are archived unchanged beside the specified contract, so both runs remain reproducible. A single evaluation script regenerates every table reported here from the raw records through the --unit-mode and --rank-key-mode switches, writing each scoring to its own directory; an alignment audit script verifies for each of the 30 tasks that the live Eurostat slice returned during execution equalled the frozen snapshot. Paired comparisons include task-cluster bootstrap intervals from 50,000 resamples with seed 20260831. No model output was altered, and no run was repeated to obtain a better score.